\documentclass{bmvc2k}

\usepackage{times}
\usepackage{epsfig}
\usepackage{graphicx}
\usepackage{amsmath}
\usepackage{amssymb}
\usepackage{multirow}
\usepackage{booktabs}
\usepackage{color}
\usepackage{footnote}
\usepackage{engord}

\usepackage{marvosym}
\usepackage{lipsum}

\usepackage{appendix}

\newtheorem{definition}{Definition}

\newcommand{\revise}[1]{{\color{red}#1}}

\newcommand{\comment}[1]{}

\newcommand{\executeiffilenewer}[3]{%
	\ifnum\pdfstrcmp{\pdffilemoddate{#1}}%
	{\pdffilemoddate{#2}}>0%
	{\immediate\write18{#3}}\fi%
}

\graphicspath{{./}{section/figure/}{images/}}

\title{Practical Transferability Estimation for Image Classification Tasks}

\addauthor{Yang Tan}{tany19@mails.tsinghua.edu.cn}{0}
\addauthor{Yang Li \textsuperscript{\Letter}}{yangli@sz.tsinghua.edu.cn}{0}
\addauthor{Shao-Lun Huang}{shaolun.huang@sz.tsinghua.edu.cn}{0}

\addinstitution{
Tsinghua-Berkeley Shenzhen Institute\\
 Tsinghua University, China
}

\runninghead{Y. Tan, Y. Li, S. Huang}{Practical Transferability Estimation}

\def\etal{\emph{et al}\bmvaOneDot}

\begin{document}

\maketitle

\revise{This paper is not the latest version. Please refer to \href{https://ieeexplore.ieee.org/abstract/document/10420486}{\textit{Transferability-Guided Cross-Domain Cross-Task Transfer Learning (IEEE TNNLS'24)}} for more details.}

\begin{abstract}

Transferability estimation is an essential problem in transfer learning to predict how good the performance is when transferring a source model (or source task) to a target task. Recent analytical transferability metrics have been widely used for source model selection and multi-task learning. A major challenge  is how to make transfereability estimation robust under the  cross-domain cross-task settings. The recently proposed OTCE score solves this problem by considering both domain and task differences, with the help of transfer experiences on auxiliary tasks, which causes an efficiency overhead. In this work, we propose a practical transferability metric called JC-NCE score that dramatically improves the robustness of the task difference estimation in OTCE, thus removing the need for auxiliary tasks. Specifically, we build the joint correspondences between source and target data via solving an optimal transport problem with a ground cost considering both the sample distance and label distance, and then compute the transferability score as the negative conditional entropy of the matched labels.   Extensive validations under the intra-dataset and inter-dataset transfer settings demonstrate that our JC-NCE score outperforms the auxiliary-task free version of OTCE  for 7\% and 12\%, respectively, and is also more  robust than other existing transferability metrics on average.

\end{abstract}

\section{Introduction}
\label{sec:intro} 
\begin{figure}[t]
    \centering
    \includegraphics[width=0.7\linewidth]{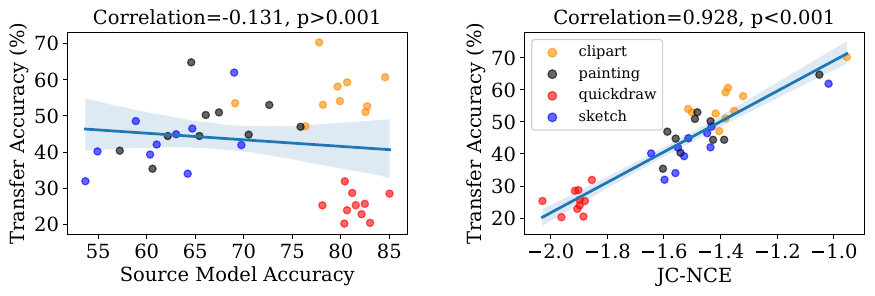}
    \vspace{-0.3cm}
    \caption{Transfer 40 source models (randomly generated 50-categories classification tasks, corresponding to each point in the figure) from \textit{Clipart, Painting, Quickdraw, Sketch} domains to a target task (25-categories) in \textit{Real} domain, which demonstrates that it is unreliable to perform source model selection according to the source model accuracy, but our JC-NCE score can predict the transfer performance more accurately. }
    \label{fig:src_acc_vs_transfer_acc}
    \vspace{-0.5cm}
\end{figure}

Transferring a related pretrained source model to a new target task usually achieves higher performance than training from scratch on target data, especially when there are only few labeled target data for supervision~\cite{pratt1993discriminability, sun2019meta}. A common pitfall in selecting which source model to transfer is basing the selection on the source accuracy. In fact, higher source model accuracy does not always lead to higher transfer accuracy due to the non-trivial differences between source and target tasks, as shown in Figure \ref{fig:src_acc_vs_transfer_acc}. Therefore, understanding the relationship between source and target tasks is crucial to the success of transfer learning. Transferability characterizes such relationship via quantitatively evaluating how easy it is to transfer the knowledge learned from a source task to the target task. In practical scenarios~\cite{NCE,bao2019information,LEEP,OTCE}, we can apply a transferability metric to directly select the best source model for a target task rather than trying each source model on the target data, which involves expensive computation. In addition, transferability can help prioritize different tasks for joint training~\cite{zamir2018taskonomy} and multi-source feature fusion~\cite{OTCE}. 

Although theoretical analyses~\cite{ben2010theory, ben2003exploiting, blitzer2008learning, mansour2009domain} in generalization bounds have suggested that the transfer performance could be attributed to several factors, e.g., certain divergence between source and target distributions, it is difficult to accurately estimate each factor from limited practical data. Meanwhile, previous empirical transferability metrics~\cite{zamir2018taskonomy, yosinski2014transferable, achille2019task2vec} suffer heavy computation burdens in retraining the source model to obtain the training loss or validation accuracy for indicating transferability. Recent analytical transferability metrics~\cite{NCE,bao2019information,LEEP,OTCE} are evidently more efficient to compute from practical data, but there also exists some drawbacks, e.g., strict data assumptions~\cite{NCE,bao2019information}, insufficient performance~\cite{LEEP}. And the state-of-the-art method OTCE~\cite{OTCE} requires auxiliary tasks with known transfer accuracy for calculating the coefficients of a linear model, which involves extra computations and restricts its application scenarios. \par

Consequently, Tan \etal~\cite{OTCE} also propose a simplified version of OTCE, namely \textbf{OT-based NCE} score, that does not depend on auxiliary tasks. It builds a soft correspondence between source and target data via solving an Optimal Transport (OT)~\cite{kantorovich1942translocation} problem, and then use the Negative Conditional Entropy (NCE) between the coupled source and target labels to characterize transferability. However, their correspondences only depend on the marginal distribution of input samples, without considering the label information. While it still outperforms previous auxiliary-task free metrics including NCE~\cite{NCE}, H-score~\cite{bao2019information} and LEEP~\cite{LEEP}, it is more reasonable to utilize both the sample information and the label information to build the joint correspondences between source and target datasets. \par 

Motivated by this idea, we propose the \textbf{JC-NCE} (Joint Correspondences Negative Conditional Entropy) score to further improve the transferability estimation performance. Inspired by recent OTDD~\cite{OTDD} method, we define the ground cost metric in the OT problem as a weighted combination of the sample distance and the label distance. By solving the OT problem, we can obtain the joint probability distribution of source and target data and then compute our JC-NCE score as the negative conditional entropy. We conduct extensive cross-domain cross-task transfer experiments to validate the superior performance of our JC-NCE score. Specifically, we first follow the same \textit{intra-dataset} experimental settings as the OT-based NCE score, i.e., perform transfer learning on two cross-domain datasets DomainNet~\cite{DomainNet} and Office31~\cite{Office31}. Results show that our JC-NCE score outperforms the OT-based NCE score with 7\% gain on average. Moreover, we conduct the \textit{inter-dataset} evaluation. We select 15 source models and 7 target datasets from the VTAB~\cite{VTAB} benchmark to perform cross-dataset transfer. Results also show that our method outperforms the OT-based NCE score with about 12\% gain. In addition, we analyze the effect of hyper parameter and compare the computation efficiency among existing metrics.\par

 In summary, our main contribution is proposing a practical transferability metric JC-NCE score which is easier to use and more efficient than the state-of-the-art OTCE score and more accurate than the simplified version OT-based NCE score with up to $12\%$ gain.

\section{Related Works}
\label{sec:relatedwork}
Theoretical analyses~\cite{maurer2009transfer,ben2010theory,ben2003exploiting,ben2006analysis,blitzer2008learning,mansour2009domain}  of generalization bounds have summarized several factors affecting the transfer performance, which also inspires the study in transferability estimation. For instance, Ben-David \etal~\cite{ben2006analysis, ben2010theory} attribute the transfer performance to the empirical risk of source task, the distance between source and target data, and the discrepancy of labeling functions. However, it is difficult to verify whether the assumptions of these theoretical works are satisfied on practical data and even more difficult to compute exactly. \par

Several empirical transferability estimation methods~\cite{zamir2018taskonomy, achille2019task2vec, ying2018transfer} are proposed to deal with practical tasks. Taskonomy~\cite{zamir2018taskonomy} propose a transferability score named \textit{task affinity} which is computed by retraining the source model on target tasks and then evaluating the transfer performance. Task2Vec~\cite{achille2019task2vec} retrains a large scale probe neural network on target tasks and then compute the Fisher information matrix to produce embedding vectors. Measuring the distance between vectors will indicate the transferability. Ying \etal~\cite{ying2018transfer} propose to learn previous transfer skills for future target tasks. Generally, empirical methods usually require heavy computation in retraining neural network, which is not superior to directly using the empirical risk of the retrained source model on  target tasks. \par

Recent analytical transferability metrics~\cite{NCE,LEEP,bao2019information, OTCE} mostly avoid the expensive computation for retraining the source model and can efficiently estimate the transferability, which is useful in source model selection. However, they still have limitations. NCE~\cite{NCE} assumes both the source and target tasks are defined on the same data instances. H-score~\cite{bao2019information} assumes the same data distribution of the source and target tasks. LEEP~\cite{LEEP} does not work sufficiently well under the challenging cross-domain cross-task transfer settings. Although OTCE~\cite{OTCE} achieves the state-of-the-art performance, it requires several auxiliary tasks with known transfer accuracy for determining the linear combination of the domain difference and the task difference, which brings extra computation and is not achievable in some scenarios. Alternatively, Tan \etal~\cite{OTCE} also propose a simplified version of the OTCE score, namely OT-based NCE score, which still can be further improved. 
\section{Method}
\label{sec:method}

In this section, we first present the definition of tranferability for classification tasks, and then introduce the main concepts of previous OT-based NCE score~\cite{OTCE}. Then we propose our JC-NCE score. 

\subsection{Transferability Definition}
Formally, we have source data $D_s = \{(x^i_s, y^i_s)\}_{i=1}^m \sim P_s(x,y)$ and target data $D_t = \{(x^i_t, y^i_t)\}_{i=1}^n \sim P_t(x,y)$, where $x^i_s, x^i_t \in \mathcal{X}$ and $ y^i_s \in \mathcal{Y}_s, y^i_t \in \mathcal{Y}_t$. Meanwhile, $P(x_s) \neq P(x_t)$ and $\mathcal{Y}_s \neq \mathcal{Y}_t$ indicate different domains and tasks respectively. In addition, we are given a source model $(\theta_s, h_s)$ pretrained on source data $D_s$, in which $\theta_s: \mathcal{X}\rightarrow\mathbb{R}^d$ represents a feature extractor producing $d$-dimensional features and $h_s:\mathbb{R}^d \rightarrow \mathcal{P}(\mathcal{Y}_s)$ is the head classifier predicting the final probability distribution of labels, where $\mathcal{P}(\mathcal{Y}_s)$ is the space of all probability distributions over $\mathcal{Y}_s$.\par

For neural network based transfer learning, there are two representative paradigms~\cite{zhang2020overcoming}, i.e., \textit{Retrain head}~\cite{donahue2014decaf} and \textit{Finetune}~\cite{girshick2014rich}. The \textit{Retrained head} method keeps the weights of source feature extractor $\theta_s$ frozen and retrains a new head classifier $h_t$. But the \textit{Finetune} method updates the source feature extractor and the head classifier simultaneously to obtain new ($\theta_t, h_t$). Compared to \textit{Retrain head}, \textit{Finetune} trade-offs transfer efficiency for better transfer accuracy and it requires more target data to avoid overfitting~\cite{OTCE}. Usually, we choose \textit{Retrain head} when there are only few labeled target data. \par 

To obtain the empirical transferability, we need to retrain the source model via Retrain head or Finetune on target data and then evaluate the expected log-likelihood on its testing set. Formally, the empirical transferability is defined as:
\begin{definition}
	The empirical transferability from source task $S$ to target task $T$ is measured by the expected log-likelihood of the retrained $(\theta_s, h_t)$ or $(\theta_t, h_t)$ on the testing set of target task:
	\begin{equation}
		\mathrm{Trf}(S \rightarrow T) = 
		\begin{cases}
		\mathbb{E} \left[ \mathrm{log}\ P(y_t|x_t; \theta_s, h_t) \right] & \text{(Retrain head)}\\
		\mathbb{E} \left[ \mathrm{log}\ P(y_t|x_t; \theta_t, h_t) \right] & \text{(Finetune)}\\
        \end{cases},
		\label{eq:true-transferability}
	\end{equation}
which indicates how good the transfer performance is on target task $T$. \cite{NCE, OTCE} 
\end{definition}\par
Although the empirical transferability can be the golden standard of describing how easy it is to transfer the knowledge learned from a source task to a target task, it is computationally expensive to obtain. Analytical transferability metric is a function of the source and target data that efficiently approximates the empirical transferability, i.e., the \textit{ground-truth} of the transfer performance on target tasks.

\subsection{Preliminary of OT-based NCE Score~\cite{OTCE}}

Before detailing our proposed JC-NCE score, we briefly introduce the main concepts of previous OT-based NCE score to facilitate the context. Tan \etal~\cite{OTCE} propose a unified framework named \textbf{OTCE}, which characterize the \textit{domain difference} and the \textit{task difference} between source and target tasks, and use the linear combination of domain difference and task difference to describe transferability. Specifically, the OTCE score first estimates the joint probability distribution $\hat{P}(X_s, X_t)$ of source and target input instances via solving an Optimal Transport (OT) problem~\cite{kantorovich1942translocation}, which also produces the Wasserstein distance (domain difference). Then based on $\hat{P}(X_s, X_t)$, we can obtain $\hat{P}(Y_s, Y_t)$ and $\hat{P}(Y_s)$ for calculating the Conditional Entropy $H(Y_t|Y_s)$ (task difference).\par 

However, although the OTCE score shows high correlation with the transfer accuracy, it requires several auxiliary tasks (at least 3) with known transfer accuracy to learn the coefficients of the linear combination under a specified transfer configuration, which involves expensive computation in obtaining the transfer accuracy of auxiliary tasks. Moreover, the learned coefficients cannot generalize to other configurations due to the variations of data and source models. To omit the learning process, they also propose an alternative efficient implementation named \textbf{OT-based NCE} score which only uses the \textit{task difference} to characterize transferability. In other words, the OT-based NCE score trade-offs accuracy for a simpler and more efficient transferability estimation. 

\subsection{JC-NCE Score}

\begin{figure}[t]
    \centering
    \includegraphics[width=1.0\linewidth]{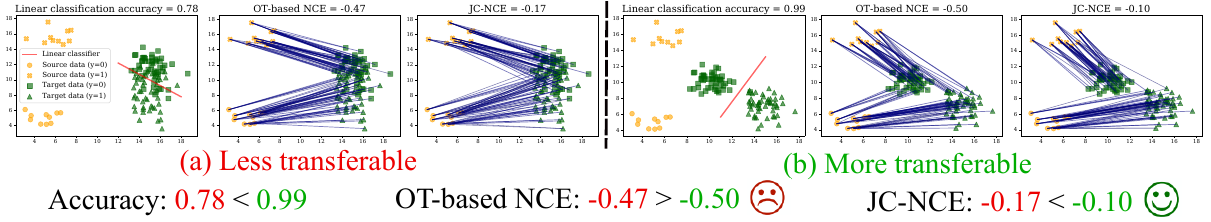}
    \vspace{-0.6cm}
    \caption{A toy example visualizes the optimal coupling between the source and target data. We can see that our JC-NCE produces a more reasonable coupling result, which ensures a better label-to-label matching than the OT-based NCE.}
    \label{fig:toy example}
    \vspace{-0.4cm}
\end{figure}

Here we propose the JC-NCE score which not only preserves the simplicity and efficiency as the OT-based NCE score but also shows higher transferability estimation performance. We also follow the framework proposed by the OT-based NCE score, i.e., build the correspondences between source and target data, and then compute the negative conditional entropy $-H(Y_t|Y_s)$ for describing transferability.\par

We adopt the ground cost metric proposed by recent OTDD~\cite{OTDD} method for building the joint correspondences, which is a weighted combination of the sample distance and the label distance. Specifically, the computation process of our JC-NCE score is described as follows. \par

First, we define the sample instances of source and target tasks as $z_s=(x_s,y_s)$ and $z_t=(x_t,y_t)$ respectively, where $z_s \in \mathcal{Z}_s = \mathcal{X} \times \mathcal{Y}_s$ and $z_t \in \mathcal{Z}_t = \mathcal{X} \times \mathcal{Y}_t$. And we define the $\alpha_y \triangleq P(X | Y=y)$, which can be estimated from a collection of finite samples with label $y$. Then the cost function can be defined as:

\begin{equation}
	d(z_s, z_t) \triangleq \lambda c(\theta_s(x_s),\theta_s(x_t)) + (1-\lambda)W(\alpha_{y_s}, \alpha_{y_t}),
\label{eq: ground distance}
\end{equation}
where $c(\cdot, \cdot) = \| \cdot - \cdot \| ^2_2$ is  the cost metric of sample distance. And $W(\alpha_{y_s}, \alpha_{y_t})$ is the 1-Wasserstein distance between labels, where the cost metric is also $c(\cdot, \cdot)$. $\lambda \in [0,1]$ is a hyper parameter to combine the sample distance and the label distance, and here we let $\lambda=0.5$. More discussion about $\lambda$ is described in Section \ref{subsec: Analysis of Combining Sample Distance and Label Distance}. It has been shown in \cite{OTDD} that Equation (\ref{eq: ground distance}) is a proper metric and a good choice for the ground cost in defining the optimal transport problem between two joint distributions $P(z_s)$ and $P(z_t)$.   
\par

Consequently, the OT problem is defined as:
\begin{equation}
	\begin{aligned}
		OT(D_s, D_t)  \triangleq  \mathop{\min}\limits_{\pi \in \Pi(D_s, D_t)}  \sum_{i,j=1}^{m,n}  d(z^i_s, z^j_t)\pi_{ij}, 
	\end{aligned}
	\label{eq: OT definition}
\end{equation}
where $\pi$ is the coupling matrix of size $m \times n$, representing the correspondences between source and target data. After solving this OT problem\footnote{The OT problem can be efficiently solved by the POT library: \url{https://pythonot.github.io}}, we obtain the optimal coupling matrix $\pi^*$. Then the empirical joint probability distribution of source and target labels, and the marginal probability distribution of source label can be easily computed as below:
   \begin{equation}
  	\hat{P}(y_s,y_t) = \sum_{i,j: y^i_s=y_s, y^j_t=y_t} \pi_{ij}^*, \ \ \hat{P}(y_s) = \sum_{y_t \in \mathcal{Y}_t} \hat{P}(y_s,y_t).
  \end{equation}
Then we can compute the JC-NCE score as the negative conditional entropy,
 \begin{equation}
	\text{JC-NCE} = - H(Y_t|Y_s) = \sum_{y_t \in \mathcal{Y}_t} \sum_{y_s \in \mathcal{Y}_s} \hat{P}(y_s,y_t)\log \frac{\hat{P}(y_s,y_t)}{\hat{P}(y_s)}.
	\label{eq: task difference}
  \end{equation}

Previous work NCE~\cite{NCE} has shown that the empirical transferability is lower bounded by the negative conditional entropy, 
\begin{equation}
	\widetilde{\mathrm{Trf}}(S \rightarrow T) \ge l_S(\theta_s, h_s) - H(Y_t|Y_s),
	\label{eq: transferability inequality}
\end{equation} 
where the training log-likelihood $\widetilde{\mathrm{Trf}}(S \rightarrow T)  = l_T(\theta_s, h_t) =  \frac{1}{n}\sum^n_{i=1}\log P(y^i_t|x^i_t; \theta_s,h_t)$ is an approximation of the empirical transferability when the retrained model is not overfitted. And $l_S(\theta_s, h_s)$ is a constant, so the empirical transferability can be attributed to the conditional entropy.\par

We show a toy example in Figure \ref{fig:toy example} to compare the optimal coupling results of the OT-based NCE score and our JC-NCE score. It can be seen that our JC-NCE score produces a more reasonable coupling between source and target data, i.e., ensure a better label-to-label matching result which leads to a more robust estimation targeting to the classification accuracy. 
\section{Experiments}
\label{sec:experiment}

We conduct extensive cross-domain cross-task transfer learning experiments to evaluate the effectiveness of our proposed JC-NCE score. First, we investigate the performance under the \textit{intra-dataset} transfer setting, i.e., source task and target task are generated from the same dataset but different sub-domains. We adopt the largest-to-date cross-domain dataset DomainNet~\cite{DomainNet} and the popular Office31~\cite{Office31} dataset. Furthermore, we study the \textit{inter-dataset} transfer setting, i.e., source task and target task are defined on different datasets. We follow the configurations of VTAB~\cite{VTAB}, a large-scale visual task adaptation benchmark. Finally, we make some analysis on the hyper parameter $\lambda$ and the computation efficiency.

\subsection{Evaluation on \textit{Intra-Dataset} Transfer Setting}
\label{subsec: evaluation on intra-dataset transfer setting}
Tasks in this setting are generated by sampling different  sets of categories from two popular cross-domain datasets including:
\begin{itemize}
	\item \textbf{DomainNet~\cite{DomainNet}} contains images distributing in six domains (styles) including \textit{Clipart (C), Infograph (I), Painting (P), Quickdraw (Q), Real (R)} and \textit{Sketch (S)}. Each domain covers 345 common object categories. Following the experimental configuration of~\cite{OTCE}, we exclude \textit{Infograph} for its noisy annotations and restrict the number of instances per category to be at most 100. 
	
	\item \textbf{Office31~\cite{Office31}}  is a representative benchmark dataset in transfer learning area. It contains 4,110 images distributing in three domains, i.e., \textit{Amazon (A), DSLR (D)} and \textit{Webcam (W)}. Each domain covers 31 categories typically found in office environment.  
	
\end{itemize}

For fair comparison, we follow the same experimental set-ups in OT-based NCE~\cite{OTCE}, i.e., the \textit{standard} configuration in which tasks have different category size, and the more challenging \textit{fixed category size} configuration. We also use Pearson correlation coefficient like~\cite{OTCE,LEEP,NCE} to evaluate the correlation between the transfer accuracy and the transferability score. We train eight ResNet-18~\cite{he2016deep} neural networks (5 for DomainNet, 3 for Office31)  as source models for each domain targeting to  the randomly generated source tasks. Specifically, the source task for DomainNet is a randomly sampled 44-category classification task, and the source task for Office31 is a 15-category classification task.\par

For \textit{standard} evaluation, we conduct 2,000 ($5\times4\times100$) cross-domain cross-task transfer tests on DomainNet, and 600 ($3\times2\times100$) tests on Office31. Specifically, we successively take one domain as the source domain, and rests are target domains. For each target domain, we randomly sample 100 classification tasks where the number of categories range from 10 to 100 for DomainNet, and 10 to 31 for Office31. The transfer accuracy on target task is the testing accuracy after retraining the head classifier of source model on target data with SGD optimizer and cross-entropy loss for 100 epochs. \par

The \textit{Fixed category size} evaluation~\cite{OTCE} is a more challenging configuration since it requires the transferability score to capture the more subtle variations of domain and the task relatedness except for the intrinsic complexity of the target task. Thus in this configuration, we randomly sample 100 target tasks with 50 categories for each target domain to keep similar task complexities, and other settings are the same as the \textit{standard} evaluation.\par

\begin{figure*}[h]
\centering
\def\svgwidth{1.0\textwidth}
	\executeiffilenewer{images/fixed_category_setting_comparison.svg}{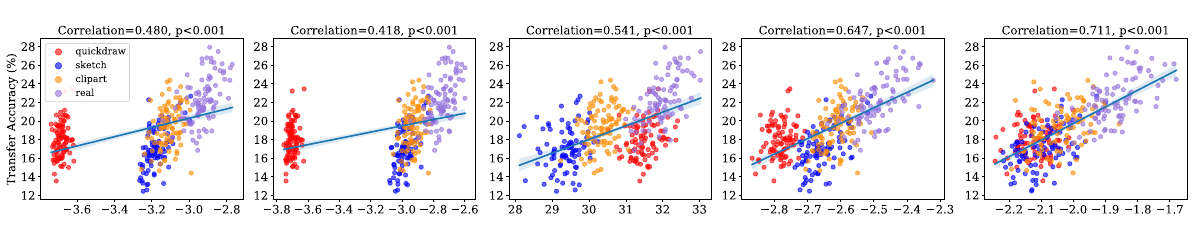}%
	{inkscape -z -D --file=images/fixed_category_setting_comparison.svg %
		--export-pdf=images/fixed_category_setting_comparison.pdf --export-latex}%
\begingroup%
  \makeatletter%
  \providecommand\color[2][]{%
    \errmessage{(Inkscape) Color is used for the text in Inkscape, but the package 'color.sty' is not loaded}%
    \renewcommand\color[2][]{}%
  }%
  \providecommand\transparent[1]{%
    \errmessage{(Inkscape) Transparency is used (non-zero) for the text in Inkscape, but the package 'transparent.sty' is not loaded}%
    \renewcommand\transparent[1]{}%
  }%
  \providecommand\rotatebox[2]{#2}%
  \newcommand*\fsize{\dimexpr\f@size pt\relax}%
  \newcommand*\lineheight[1]{\fontsize{\fsize}{#1\fsize}\selectfont}%
  \ifx\svgwidth\undefined%
    \setlength{\unitlength}{575.43304924bp}%
    \ifx\svgscale\undefined%
      \relax%
    \else%
      \setlength{\unitlength}{\unitlength * \real{\svgscale}}%
    \fi%
  \else%
    \setlength{\unitlength}{\svgwidth}%
  \fi%
  \global\let\svgwidth\undefined%
  \global\let\svgscale\undefined%
  \makeatother%
  \begin{picture}(1,0.20197045)%
    \lineheight{1}%
    \setlength\tabcolsep{0pt}%
    \put(0,0){\includegraphics[width=\unitlength,page=1]{fixed_category_setting_comparison.pdf}}%
    \put(0.27,0.18995854){\makebox(0,0)[lt]{\lineheight{1.64999998}\smash{\begin{tabular}[t]{l}\footnotesize{Fixed category size setting, source domain: Painting}\end{tabular}}}}%
    \put(0.08,0.0){\makebox(0,0)[lt]{\lineheight{1.64999998}\smash{\begin{tabular}[t]{l}\footnotesize{LEEP\cite{LEEP}}\end{tabular}}}}%
    \put(0.28,0.0){\makebox(0,0)[lt]{\lineheight{1.64999998}\smash{\begin{tabular}[t]{l}\footnotesize{NCE\cite{NCE}}\end{tabular}}}}%
    \put(0.465,0.0){\makebox(0,0)[lt]{\lineheight{1.64999998}\smash{\begin{tabular}[t]{l}\footnotesize{H-score\cite{bao2019information}}\end{tabular}}}}%
    \put(0.62,0.0){\makebox(0,0)[lt]{\lineheight{1.64999998}\smash{\begin{tabular}[t]{l}\footnotesize{OT-based NCE\cite{OTCE}}\end{tabular}}}}%
    \put(0.87,0.0){\makebox(0,0)[lt]{\lineheight{1.64999998}\smash{\begin{tabular}[t]{l}\footnotesize{JC-NCE}\end{tabular}}}}%
    \put(0,0){\includegraphics[width=\unitlength,page=2]{fixed_category_setting_comparison.pdf}}%
  \end{picture}%
\endgroup%
%

\caption{Visualization of the correlations between the transfer accuracy and transferability scores under the challenging \textit{fixed category size setting}, where all target tasks (50-categories classification, corresponding to each point in the figure) have similar complexities. Our JC-NCE score significantly outperforms the OT-based NCE score, especially as illustrated in the \textcolor{green}{green circle}.}

\label{fig: correlation fixed_category_setting_comparison}
\vspace{-0.2cm}
\end{figure*}

Table \ref{tab: intradataset comparsion} shows the comparisons among our JC-NCE score and other analytical transferability metrics, including the OT-based NCE, LEEP, NCE and H-score in both experimental configurations. The average correlation scores of JC-NCE are 0.914 and 0.615 respectively, which significantly outperforms the compared methods. In particular, under the fixed category size configuration, the JC-NCE score achieves a 13\% improvement compared to the state-of-the-art OT-based NCE score. This improvement can be visually captured in Figure \ref{fig: correlation fixed_category_setting_comparison}, where the transferability scores of target tasks in domain \textit{Quickdraw} (in red) can be better estimated via the JC-NCE score. More visual comparisons are shown in the Supplementary.

\begin{table*}[t]
\begin{minipage}{\linewidth}

\setlength\tabcolsep{3pt} 

\footnotesize

\caption{Quantitative comparisons evaluated by Pearson correlation coefficients between the transfer accuracy and transferability scores under the \textbf{intra-dataset} transfer setting. }
\label{tab: intradataset comparsion}

\centering

\begin{tabular}{cccc cccc}
\toprule

\multirow{2}{*}{Config} & Source& Target& \multirow{2}{*}{JC-NCE }& \multirow{2}{*}{OT-based NCE\cite{OTCE}} & \multirow{2}{*}{LEEP\cite{LEEP}} & \multirow{2}{*}{NCE\cite{NCE}} & \multirow{2}{*}{H-score\cite{bao2019information}}  \\
& domain & domain & & & & &\\

\midrule

\multirow{9}{*}{Standard} & C & P,Q,R,S & \underline{0.952} & \textbf{0.960} & 0.919 & 0.787 & -0.864 \\
															& P & C,Q,R,S & \textbf{0.953} & \underline{0.952} & 0.886 & 0.812 & -0.858 \\
															& Q & C,P,R,S & \textbf{0.968} & \underline{0.963} & 0.942 & 0.935 & -0.843 \\
															& R& C,P,Q,S & \textbf{0.957} & \underline{0.951} & 0.892 & 0.851 & -0.870 \\
															& S & C,P,Q,R& 0.951 & \textbf{0.959} & 0.952 & \underline{0.954 }& -0.882 \\
															& A & D,W      & \textbf{0.817} & \underline{0.813} & 0.805 & 0.796 & -0.590 \\
															& D & A,W     & \textbf{0.867} & 0.843 & \underline{0.857} & 0.849 & -0.441 \\
															& W & A,D     & \textbf{0.845} & 0.803 & \underline{0.811} & 0.804 & -0.489 \\
\cmidrule{2-8} 
													  & &Average & \textbf{0.914} & \underline{0.906} & 0.883 & 0.849 & -0.730 \\

\midrule
\midrule

									& C & P,Q,R,S &  \textbf{0.754}& \underline{0.729} & 0.614 & 0.535 & 0.599 \\
									& P & C,Q,R,S & \textbf{0.711}  & \underline{0.647} & 0.480 & 0.418 & 0.541 \\
Fixed category size& Q & C,P,R,S & \textbf{0.427} & \underline{0.306}  & 0.213 & 0.269 & 0.288 \\
			    	& R& C,P,Q,S& \textbf{0.677} & \underline{0.587}  & 0.465 & 0.440 & 0.100* \\
								&    S & C,P,Q,R& \textbf{0.506}  & \underline{0.443} & 0.381 & 0.427 & 0.302 \\
\cmidrule{2-8} 
& &Average & \textbf{0.615} & \underline{0.542} & 0.431 & 0.418 & 0.366 \\

\bottomrule

\end{tabular}
\footnotesize{Superscript $^*$ denotes $p>0.001$, and \textbf{bold} denotes the best result, and \underline{underline} denotes the \engordnumber{2} best result.}
\end{minipage}
\end{table*}

\begin{table*}[t]
\begin{minipage}{\linewidth}

\setlength\tabcolsep{3pt} 

\footnotesize

\caption{Quantitative comparisons evaluated by Pearson correlation coefficients between the transfer accuracy and transferability scores under the \textbf{inter-dataset} transfer setting. The upper part represents transferring via Finetune, and the lower part represents Retrain head.}
\label{tab: interdataset comparsion}

\centering

\begin{tabular}{l | ccccccc |c}
\toprule

Method & Caltech101 & CIFAR-100 & DTD & Flowers102 & Pets & Camelyon & SVHN &   Avg\\
\midrule
JC-NCE & \textbf{0.784} & \underline{0.938} & \underline{0.905} & \underline{0.973} & \underline{0.915} & \underline{0.646*} & \textbf{0.670* }& \underline{0.833} \\
OT-based NCE\cite{OTCE} & \underline{0.685*} & 0.764 & 0.819  & 0.779  & 0.818 & 0.494* & 0.592* & 0.707 \\
H-score\cite{bao2019information} &0.680* & \textbf{0.957} & \textbf{0.970} & \textbf{0.991} & \textbf{0.980} & \textbf{0.693*} & \underline{0.666*} & \textbf{0.848}\\

\midrule
\midrule

JC-NCE & \underline{0.935} & \textbf{0.939} & \underline{0.903} & \underline{0.919} & \underline{0.963} & 0.710* & \underline{0.887} & \underline{0.894} \\
OT-based NCE\cite{OTCE} & 0.891 & \underline{0.906} & 0.869 & 0.856 & \textbf{0.981} & \underline{0.735*} & 0.686* & 0.846 \\
H-score\cite{bao2019information} & \textbf{0.983} & 0.879 & \textbf{0.952} & \textbf{0.973} & 0.877 & \textbf{0.898} & \textbf{0.932} & \textbf{0.928}\\

\bottomrule

\end{tabular}

\footnotesize{Superscript $^*$ denotes $p>0.001$, and \textbf{bold} denotes the best result, and \underline{underline} denotes the \engordnumber{2} best result.}

\end{minipage}

\end{table*}

\subsection{Evaluation on \textit{Inter-Dataset} Transfer Setting}
\label{subsec: evaluation on inter-dataset transfer setting}

We further study the performance under the \textit{inter-dataset} transfer setting, where source models are provided by the Visual Task Adaptation Benchmark (VTAB)~\cite{VTAB}. The model zoo contains 15 models trained on ImageNet by different algorithms, e.g., supervised learning (Sup-100\%) , semi-supervised learning (Semi-rotation-10\% and Semi-exemplar-10\%~\cite{zhai2019s4l}) , self-supervised learning (Rotation~\cite{gidaris2018unsupervised} and Jigsaw~\cite{noroozi2016unsupervised}), generative method (Cond-biggan~\cite{brock2018large}) and VAEs~\cite{kingma2013auto}, etc. For target tasks, we introduce 7 image classification datasets including Caltech101~\cite{caltech101}, CIFAR-100~\cite{cifar100}, DTD~\cite{dtd}, Flowers102~\cite{flowers102}, Pets~\cite{pets}, SVHN~\cite{svhn} and Camelyon~\cite{camelyon}. More information about source models and target datasets is described in the Supplementary. \par

Specifically, we transfer source models to each target task via two transfer methods, i.e., \textit{Retrain head} (only retrain a new head classifier) and \textit{Finetune} (finetune all weights). We follow the transfer accuracy reported in VTAB. As the VTAB model zoo only publicly provides feature extractors, we are unable to make comparisons with the LEEP and NCE scores since they require the entire source model for predicting the pseudo labels on target data so that they cannot be used for transferring feature extractor only.\par

We show the correlation comparisons among our JC-NCE, OT-based NCE  and H-score  in Table \ref{tab: interdataset comparsion}. Our JC-NCE score also outperforms the OT-based NCE score. Note that  H-score   achieves slightly better correlation results than JC-NCE. Because the source dataset and most target datasets come from the natural environment so that the domain gap is small, which satisfies the data assumption of H-score. However, in the previous \textit{intra-dataset} experiment (Table \ref{tab: intradataset comparsion}), H-score is negatively correlated with the transfer performance, failing to estimate the cross-domain transferability score. Therefore, we conclude that JC-NCE is a more robust and practical metric overall. Visual comparisons are included in the Supplementary.\par

We also make comparisons in source model selection shown in Table \ref{tab: model selection}. Each target task has 15 candidate source models and we want to verify whether the source model with the highest transferability score is the best source model (with highest transfer accuracy). We calculate the Top-k (k=1,2,3) selecting accuracy and found that the JC-NCE, OT-based NCE and H-score achieved comparable good results, i.e., the ground-truth best model can be selected from the predicted Top-3 highest transferable models in most cases.   

\begin{table}[h]
\begin{minipage}{\linewidth}

\footnotesize

\caption{The Top-k accuracy of the best source model selection under the \textbf{inter-dataset} transfer setting, i.e., select the best one from 15 source models for 7 target tasks according to their transferability scores. The upper and lower parts represent transferring via Finetune and Retrain head respectively.}
\label{tab: model selection}

\centering

\begin{tabular}{l | ccc}
\toprule

Method & Top-1 & Top-2 & Top-3 \\

\midrule

JC-NCE &1 / 7 & 5 / 7 & 5 / 7 \\
OT-based NCE\cite{OTCE} & \textbf{2 / 7} & \textbf{7 / 7} &  \textbf{7 / 7}   \\
H-score\cite{bao2019information} &\textbf{2 / 7} & 5 / 7 & 5 / 7 \\

\midrule
\midrule

JC-NCE & \textbf{3 / 7} & \textbf{5 / 7} & \textbf{6 / 7}  \\
OT-based NCE\cite{OTCE} & \textbf{3 / 7} & 4 / 7 & 5 / 7  \\
H-score\cite{bao2019information} &2 / 7 & 4 / 7 &\textbf{6 / 7} \\

\bottomrule

\end{tabular}

\end{minipage}

\end{table}

\begin{figure}[!htb]
	\centering
	\begin{minipage}[h]{0.45\linewidth}
		\centering
		\includegraphics[width=1.0\linewidth]{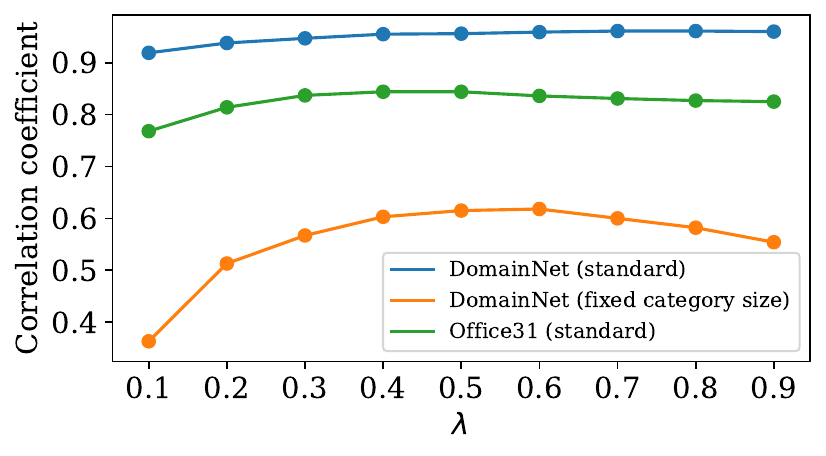}
		\vspace{-0.8cm}
		\caption{Analysis of $\lambda$.}
		\label{Fig:study of lambda}
	\end{minipage}
	\hfill
	\begin{minipage}[b]{0.45\linewidth}
		\centering
		\footnotesize
		\makeatletter\def\@captype{table}\makeatother\caption{Computation time statistics.}
		\begin{tabular}{lc}
			\toprule
			Method & Time \\
			\midrule
			Empirical transferability &   858s (14.3min)\\
			LEEP\cite{LEEP} & 0.06s \\
			NCE\cite{NCE}&  0.007s\\
			H-score\cite{bao2019information} &  0.11s\\
			OT-based NCE\cite{OTCE} &  0.41s\\
			JC-NCE &  1.87s  \\
			\bottomrule
		\end{tabular}
		\label{tab: time statistics}
	\end{minipage}
\end{figure}


\subsection{Effect of Parameter $\lambda$}
\label{subsec: Analysis of Combining Sample Distance and Label Distance}

We study the effect of the hyper parameter $\lambda \in [0,1]$ in Equation (\ref{eq: ground distance}), which determines the impacts of the sample distance and the label distance in computing the joint correspondences of source and target datasets. As shown in Figure \ref{Fig:study of lambda}, the JC-NCE score achieves the highest performance when let $\lambda=0.5$.

\subsection{Efficiency Analysis}
We compare the computation time among transferability metrics shown in Table \ref{tab: time statistics}. Specifically, the empirical transferability is computed on GPU (NIVIDIA GTX1080Ti) through retraining the source model (ResNet-18) on target data and then evaluating the log-likelihood on the testing set. For analytical metrics, we randomly sample 1,000 instances for computation (on CPU). Results demonstrate that analytical transferability metrics are evidently more efficient and easier to obtain than the empirical transferability. More implementation details are introduced in the Supplementary.

\section{Conclusion}
\label{sec:conclusion}
In this paper, we propose JC-NCE score, a practical transferability metric  for classification tasks. It preserves the simplicity and efficiency of the previous OT-based NCE method, but significantly improves its transferability estimation  performance by considering both the sample distance and the label distance simultaneously. Extensive experiments in both the intra-dataset and the inter-dataset settings demonstrate that our JC-NCE score works more robustly than previous analytical transferability metrics. In future works, we will investigate how to use JC-NCE to benefit downstream applications in heterogeneous transfer learning and multi-task learning.

\bibliography{egbib}
\end{document}